\documentclass[runningheads]{llncs}
\usepackage{graphicx}

\usepackage{natbib}
\setcitestyle{square,numbers}

\usepackage[toc,page]{appendix}

\usepackage{nameref}

\usepackage{amsmath}
\usepackage{amssymb}
\usepackage[ruled,vlined]{algorithm2e}

\usepackage{enumitem}
\usepackage{subcaption}

\usepackage{xcolor}
\usepackage{color,soul}
\usepackage{pdfpages}

\def\HiLi{\leavevmode\rlap{\hbox to \hsize{\color{yellow!50}\leaders\hrule height .8\baselineskip depth .5ex\hfill}}}
\newcommand{\newpar}[1]{\vspace{-0mm}\paragraph{#1.}}
\definecolor{myyellow}{rgb}{1.0,1.0,0.8}
         
\newcount\Comments  
\definecolor{darkgreen}{rgb}{0,0.6,0}         
         
\newcommand{\kibitz}[2]{\ifnum\Comments=1{\color{#1}{#2}}\fi}

\begin{document}

\title{Learning Aggregation Rules in Participatory Budgeting: A Data-Driven Approach}
%
%

\author{Roy Fairstein\inst{1} \and
Dan Vilenchik\inst{1} \and
Kobi Gal\inst{1,2}}
\author{Roy Fairstein\inst{1}\orcidID{0000-0002-2352-5200} \and
Dan Vilenchik\inst{1}\orcidID{0000-0002-3895-688X} \and
Kobi Gal\inst{1,2}\orcidID{0000-0001-7187-8572}}
%
%
\institute{Ben-Gurion Univ. of the Negev, Israel \and
University of Edinburgh, U.K.}
\maketitle              
\begin{abstract}
Participatory Budgeting (PB) offers a democratic process for communities to allocate public funds across various projects through voting. In practice, PB organizers  face challenges in selecting aggregation rules either because they are not familiar with the literature and the exact details of every existing rule
or because no existing rule echoes their expectations. 
This paper presents a novel data-driven approach utilizing machine learning to address this challenge. By training neural networks on PB instances, our approach learns aggregation rules that balance social welfare, representation, and other societal beneficial goals. It is able to generalize from small-scale synthetic PB examples to large, real-world PB instances. It is able to learn existing aggregation  rules but also generate new rules that adapt to diverse objectives, providing a more nuanced, compromise-driven solution for PB processes. The effectiveness of our approach is demonstrated through extensive experiments with synthetic and real-world PB data, and can expand the use and deployment of PB solutions.

\end{abstract}
%
%
%


\section{Introduction}\label{sec:intro}

Participatory budgeting (PB) is a democratic process in which participants decide how public funds should be allocated to different projects~\citep{aziz2021participatory}.  
In PB, participants vote on the projects they would like to see carried out, and an aggregation mechanism selects which projects to fund under known budget constraints.  
PB was first implemented in Porto Alegre, Brazil~\citep{cabannes2004participatory}, and its use has spread rapidly to locations such as  Germany~\citep{zepic2017participatory}, Chicago~\citep{stewart2014participatory}, New York~\citep{su2017porto} and many more.

While PB is most commonly associated with municipal budget allocations, it can be applied to a wide variety of contexts. For example, a company's human resources department might use PB to organize employee activities within a limited budget; a group of friends could use PB to decide which movies to watch or which dishes to order from takeout; or a community center might need to allocate funds to offer different activities. In each case, the desired outcome may differ: the group of friends may prioritize choosing movies that match everyone's preferences, the community center may focus on offering activities that ensure broad participation, and the HR team may aim to strike a balance between popular choices and providing something for everyone. 
This  ``desired outcome" should be formalized as an aggregation rule that  decides which options will be chosen.

A popular formalization of  an aggregation rule is an  optimization problem  over a closed-form objective function. The literature offers numerous such aggregation rules, each designed to optimize different criteria. For example, Approval Voting (AV) aims to maximize social welfare, while the Chamberlin-Courant (CC)  rule focuses on maximizing representation\cite{chamberlin1983representative}. 

In practice, PB organizers may struggle with encoding their ``desired outcome" as an explicit aggregation rule, especially if none of the existing rules in the literature suits their objectives. This can occur when PB organizers have different desired outcomes themselves (say half of the PB organizers prefer AV and half CC). In addition, objectives can shift over time, requiring to adapt and update   the aggregation rule. Without new tools to help PB organizers develop an agreeable aggregation rule, they may default to simple greedy aggregation rules whose outcomes may be suboptimal. 

An attempt to provide a framework that supports tailored rules was recently taken in the multi-winner (MW) setting by \citet{faliszewski2022optimization}. Their solution defines the desired outcome in terms of a target distribution and a matching hand-crafted scoring rule to sort the projects. However, the fact that the solution includes manually designed components limits its applicability to more complex or realistic settings where it's not always clear what the target distribution should look like.

To overcome these challenges we propose to implicitly learn aggregation rules from data. Our method uses the project data, votes, and outcomes—to embed the aggregation rule directly into a neural network (NN) through training. In this approach, the NN itself becomes the aggregation rule, eliminating the need for predefined formulas. The ``desired outcomes" are now represented by the outcomes of the  PB instances in the dataset, allowing the NN to learn aggregation rules that align with these outcomes.

For example, reconsider the PB organizing committee where half of the members prefer AV and the other half CC. By training the NN with PB instances where half of the examples are solved using AV and the other half using CC, the network learns a compromise rule that balances both objectives. This compromise is embedded within the NN and reflects a rule agreeable to the entire committee. This approach also offers a more flexible and adaptive solution that supports preferences that evolve over time by a simple fine-tuning step that can proceed whenever an update is due.

The main contributions of this paper are:  

\noindent (1) We offer an approach that learns aggregation rules directly from examples of PB votes and outcomes, enabling the discovery of arbitrarily complex aggregation rules. This removes the need to explicitly define desired properties or manually craft rules that meet them.

\vspace{1.5mm}

\noindent (2) The NN generalizes from small PB instances, with as few as 10 to 900 voters, to larger, real-world cases in the tens of thousands. This shows that our approach is practical and actionable, as users can provide simple, small-scale examples that are easily scalable to larger real-world instances.

\vspace{1.5mm}

\noindent (3) We show that using our approach, three aggregation rules widely used in the literature, AV, CC, and PAV, can be learned from data. We demonstrate this ability on several datasets with very different characteristics, including one real-world dataset. 

\vspace{1.5mm}

\noindent (4) To illustrate a scenario with multiple objectives, we developed a family of rules that interpolate between AV and CC, balancing social welfare and representation. By generating a mix of PB instances — some solved using AV and others CC— we show that the model can learn compromise rules. For instance, a 30\%-70\% blend produces a rule that closely resembles Proportional Approval Voting (PAV), offering a data-driven rationale for PAV's balance of welfare and representation. Future work will examine, through experimentation with human participants, the ability to define ad-hoc aggregation rules via examples, then learn them and evaluate their outcomes by asking humans for their opinion on the NN's decisions.

\vspace{1.5mm}

The remainder of the paper is organized as follows: In Section \ref{sec:rw}, we discuss the background and previous work related to participatory budgeting (PB), focusing on aggregation rules and their theoretical underpinnings. In Section \ref{sec:motiv}, we elaborate on the motivation for using ML thought concrete examples, and in Section \ref{sec:method} we introduce our methodology, describing the Set Transformer architecture ~\citep{anil2021learning} and its adaptation for PB, as well as the training strategies employed. Section \ref{sec:data} outlines the datasets used in our experiments, including both synthetic and real-world instances. Section \ref{sec:res1} presents the evaluation results for learning target distributions, where we analyze the model's performance across various aggregation rules. In Section \ref{sec:AV+CC}, we extend the evaluation to learning mixture distributions and discuss the model’s capacity to learn compromises between competing aggregation rules. Finally, in Section \ref{sec:conc}, we summarize the findings and suggest directions for future work.

\subsection{Notation}\label{sec:prem}
We provide basic notation  and mathematical expressions that will be used throughout the paper.

For   $k\in \mathbb{N}$,  let $[k] := \{1,\ldots,k\}$. A PB instance is defined by a set of $n$ voters $N=[n]$ that express their preferences over a set of $m$ projects $P=\{P_1,\ldots,p_m\}$, such that each project $p\in P$ has cost   $c(p)\in \mathbb{R}^+$ (in the case of multi-winner all projects have a cost of 1) and there is a total budget of $L$. We will denote the approval profile by $A:=(A_1,\ldots,A_n)$ and for any voter $i\in N$, $A_i\subseteq P$ is the set of projects that voter $i$ approves. The aggregation method chooses a subset of projects $B\subseteq P$ to fund such that its total cost does not exceed the budget (feasible bundle), $c(B) :=  \sum_{p\in B} c(p) \leq L$. 


\section{Related Work}\label{sec:rw}


Within the PB literature, our results relate to work on aggregating voters' preferences. Prior work focused on algorithms for aggregating voters' preferences to guarantee some desirable social criteria. Those guarantees can be divided into three main categories: social welfare and representation, fairness and proportionality, and a variety of axioms.

Concerning social welfare and representation (which are our primary measures of evaluation), previous works studied the bounds of existing aggregation methods and how to optimize these objectives. \citet{fairstein2022welfare}  examined the trade-off between social welfare and representation of several aggregation methods, while \citet{jain2020participatory} focused on the complexity of optimizing social welfare. Those metrics will also help us in this work for evaluation.

Another approach for aggregation rule evaluation consists of axioms such as guaranteeing that a project is funded by the rule even if lowering its cost~\citep{talmon2019framework}. In addition, much attention was given to the notion of proportionality, guaranteeing that a big group of voters who vote similarly should be guaranteed some minimal welfare.
For more detail on evaluation and PB in general, we refer the reader to the comprehensive survey by \citet{rey2023computational}.




While there is a rich literature on analyzing existing aggregation rules, relatively little work has been dedicated to the creation of new aggregation rules. A relatively new aggregation method that received a lot of attention is the Method of Equal Shares~\citep{peters2021proportional} (ES), which is known to guarantee proportional outcomes while running in polynomial time. Many works that suggest new rules are variations of ES that are applied to different PB settings, for example, when there is a mix of divisible and indivisible projects~\citep{lu2024approval}, or when there exists substitute projects~\citep{fairstein2021proportional}. Our approach directly addresses the challenge of creating new PB aggregation rules by learning rules from data.

 On the connection between voting rules and ML, \citet{procaccia2009learnability} study the PAC-learnability of single-winner voting rules, and \citet{caragiannis2022complexity} study the PAC-learnability in the multi-winner settings.
These works demonstrated that learning aggregation rules from examples is computationally hard. 
They show that while learning is feasible in the PAC learning framework, it is feasible in the worst case only in computationally inefficient ways.

Recent work has shown the potential to learn aggregation rules using machine learning, particularly in the single-winner domain, where voters rank candidates or projects, and only one is selected without budget constraints. For instance, \citet{doucette2015conventional} predicts outcomes from partial preferences, \citet{kujawska2020predicting} mimics various voting rules, and \citet{firebanks2020machine} creates a framework for planners to define voting rules based on constraints. However, these methods rely on fixed-size inputs, requiring separate networks for different problem sizes. To address this, \citet{anil2021learning} used flexible architectures like Deep-sets \citep{zaheer2017deep}, Graph Neural Networks \citep{scarselli2008graph,xu2018powerful}, and Set Transformers ~\citep{lee2019set}, enabling training on small problems and generalizing to larger ones.



We extend the Set Transformer architecture of \citet{anil2021learning} to the participatory budgeting domain, which differs from the single-winner domain in three significant aspects. First, unlike in the single-winner setting, where the voters usually report a ranking over all candidates, in the PB setting, each voter approves up to $k$ projects. Second, the outcome of a PB event includes, in most cases, more than one project. Finally, each project has a cost, and the total cost of all approved projects shouldn't exceed a predetermined budget. All these differences make the extension challenging, where handling the costs and budget correctly is the biggest challenge.





\section{Motivation}\label{sec:prob_def}\label{sec:motiv}

In this paper, we address the challenge of developing new aggregation rules tailored to specific objectives. A key consideration is enabling individuals to receive a personalized aggregation rule without requiring a deep understanding of the theoretical aspects of PB.
We use an example from the work by~\citet{faliszewski2022optimization} to demonstrate the limitations of that approach and the solution we offer. 

\citet{faliszewski2022optimization} focus on the  Euclidean domain where  voters and projects are objects in a $d$-dimensional vector space. Voters' preferences are usually a decreasing function of the distance between voters and projects. Motivated by an assumption that more people live in the city center, MW designers can define an optimal distribution whose support lies entirely on the city's center. To realize this plan, the aggregation rule should choose projects as close to the center as possible. \citet{faliszewski2022optimization} do this by manually crafting a scoring rule, using an optimization program to compute the optimal scoring weights that minimize the objective. 

Another example is a target distribution that for some pre-defined fixed $x \in \mathcal{R}$, splits the support evenly between the points $x$ and $-x$ in the 1-dimensional Euclidean domain. The objective is still simple enough, requiring to minimize the distance from those two points.

Now consider a 2-dimensional geographical area. In this case, it becomes unclear how to extend the ``twin peaks" distribution to 2D. Should it consists of four points at $x$ and $-x$ in each dimension? Or should be represented by a circle with a radius of $x$. At any case, the 1D scoring rule does not generalize easily to 2D, and coming up with a new scoring rule is challenging.

Finally, let us mention an additional complication when using the Euclidean domain in the PB domain, where project costs are introduced. Going back to the city-center rule, a single expensive project might be preferred over many cheaper projects because it is slightly closer to the center.  Thus the ranking rule should now take into account not only distances but also costs, making the entire task even more unfeasible.

\section{Methodology}\label{sec:method}
In this section we show how the downsides of the hand-crafted  approach can be ameliorated using the ML approach. The ML approach contains two key components: the train set (which will define implicitly the aggregation rule) and the architecture (which is going to learn that rule).

\newpar{Objective representation from training data}
In the previous section, we described one way to represent the objective: using the project distribution in the Euclidean domain. 
A real PB instance can rarely be defined in the Euclidean domain, and defining the distribution directly can quickly become impractical. For this reason, we will represent the objective using examples, i.e., a set of triplets: the votes, project costs, and an outcome.
Then, given enough examples, we train an ML pipeline that learns a rule that approximates the objective that the data induces.

The advantages of using data to define the objective are twofold: first, we are not limited to any structure (such as the Euclidean domain); thus, any set of PB instances can be used. Second, there is no need to explicitly define the objective, which can be complex on the one hand and might require inaccessible knowledge on the other hand.



\newpar{Concrete target objectives}
Three popular aggregation rules from the literature   will be used in this paper: Approval Voting (AV), Chamberlin–Courant (CC) \citep{chamberlin1983representative}, and Proportional Approval Voting (PAV) \citep{orsted1894oversigt}.
AV selects a feasible bundle $B\subseteq P$ that maximizes the social welfare (SW). The SW score of an approval profile $A$ with respect to a bundle $B$ is 
$SW(A,B) = \sum_{i\in V}|A_i\cap B|$. 

CC selects a feasible bundle $B\subseteq P$ that maximizes the representation score where the  representation  score of an approval profile $A$ with respect to a bundle $B$ is $REP(A,B)=\sum_{i\in V}\min(1,|A_i \cap B|)$

Finally, PAV selects a feasible bundle $B\subseteq P$ that maximizes the following score:
$$SC_{PAV}(A,B)=\sum_{i\in V}\sum_{k=1}^{|A_i\cap B|}\frac{1}{k}.$$

The PAV score compromises between social welfare and representation by gradually decreasing the added value that each voter gets for each additional funded project.

To illustrate these objectives think of a simple example with three projects $\{p_1,p_2,p_3\}$ each cost 1 dollar, with a budget of 2 dollars.  There are three voters such that the first approves all three projects, the second $\{p_1,p_2\}$, and the third only $p_1$.
If we fund   $\{p_2,p_3\}$, we will have social welfare of $2+1=3$ (two want $p_2$ and only one wants $p_3$) and representation of $1+1=2$ as the last voter does not get any of his approved projects.

\newpar{Learning method}
While using scoring rules to learn new rules has the advantage of simplicity, it is limited to learning more complex rules. For this reason, we would like to use neural networks (NN), which are known for their ability to solve complex tasks.

For this purpose, we define the NN's input as a set of approval votes~\footnote{While we focused on approval votes, our approach can handle other types of votes without any changes.} (for each voter, the set of projects they approve) represented by a binary matrix (called the vote matrix) and a vector with all of the project costs (in our case normalized, such that the budget is always one). The network's output is the projects that should be funded.


The specific NN architecture we chose is the Set Transformer, which was used successfully in the multi-winner setting. We'll highlight the changes and adaptations needed for the PB setting.


\begin{figure}
        \begin{subfigure}[b]{0.45\textwidth}
         \includegraphics[width=7.5cm]{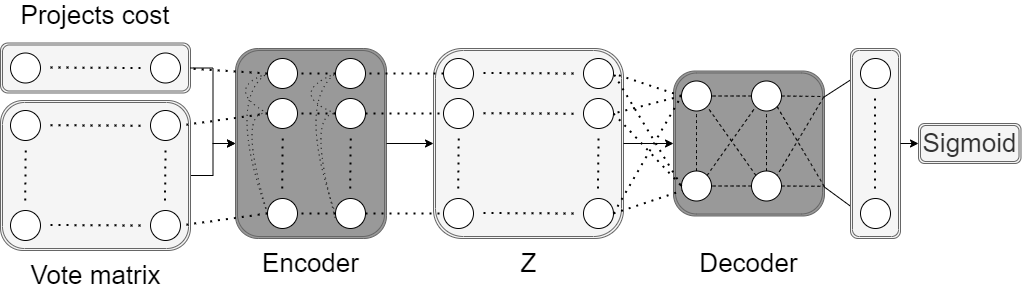}
         \caption{ST architecture, with sigmoid instead of softmax. The input dimensions are $(|N|+1) \times |P|$.}\vspace{5mm}
         \label{fig:st}
     \end{subfigure}
     \begin{subfigure}[b]{0.45\textwidth}
         \includegraphics[width=7.5cm]{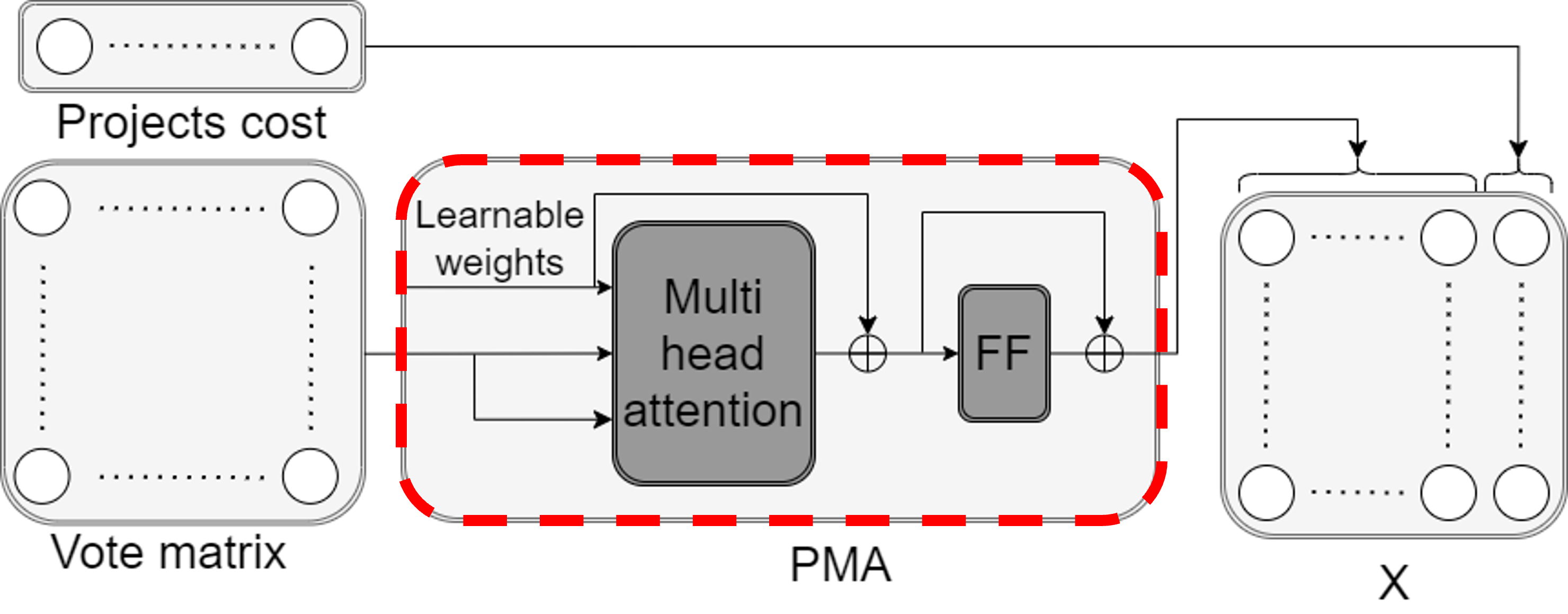}
         \caption{Adding a PMA unit before ST to embed voters matrix in the projects' latent space while eliminating voters dimension; input to ST model has now dimensions  $|P| \times c$, where $c$ is the embedding dimension.}\vspace{5mm}
         \label{fig:pma}
     \end{subfigure}
     
    \caption{The network architecture.}
\end{figure}

\subsection{Set Transformers (ST) Architecture} \label{sec:arch}
The ST network was suggested as a permutation-invariant network, i.e., the network outputs the same no matter the input order. This property is important for our scenario as we would like the outcome to be independent of the order of the voters and projects. 
For this purpose, the network combines attention blocks and linear layers to create an encoder and decoder pair; see Figure \ref{fig:st}. The encoder comprises a chain of self-attention blocks (SAB), each followed by residual feed-forward layers (rFF). 

The decoder begins with a  "Pooling by Multihead Attention" (PMA)~\citep{lee2019set}, followed by SAB and an rFF. The PMA has a similar purpose as any pooling layer - dimension reduction. This is done by using the attention layer, where we also learn the queries in addition to the regular weights. As the dimension of the outcome is defined according to the dimension of the queries, we can control the output dimension.

To accommodate PB data rather than multi-winner, the NN needs to allow project costs, and the output is not a single project but rather a bundle. Thus, we make the following changes. The input is a  matrix in  $\{0,1\}^{|N| \times |P|}$, the $i^{th}$ row corresponds to the projects approved by voter $i$. To this matrix,  another row is added with the costs of the projects normalized by the given budget so that the cost of the optimal bundle is always at most 1. Finally, the softmax at the output layer is replaced with a sigmoid function so the network outputs a distribution over the projects. The final bundle is chosen greedily by their score until exhausting the budget.~\footnote{In most cases, those projects have a value close to 1, while non-chosen projects have a value close to 0. The bundle is chosen greedily to handle the cases where the projects with a value close to one either exceed or do not exhaust the budget.}

This approach has two issues:
\begin{itemize}
    \item The ST architecture consists of an encoder responsible for generating latent space representations for voters and a decoder that utilizes PMA to translate these representations back into the project domain. However, doing so can make it more difficult for the encoder to effectively preserve and retrieve the project details within the encoded voter representations for the decoding process.
    

    \item The number of voters can grow large, unlike the number of projects, at least in a realistic setting. Therefore, the ST can only handle input matrices with a maximum of a few thousand voters before the GPU memory is exhausted. 
\end{itemize}

\newpar{ST + PMA Architecture}
To solve these two issues, we move the PMA layer from the decoder to before the encoder, as seen in Figure~\ref{fig:pma}. The PMA layer now embeds the vote matrix in the {\em projects'} latent space (rather than the voters').
The costs are then added to the PMA results and forwarded to the ST (without the PMA in the decoder). 

Before running the encoder-decoder, this procedure changes our input dimensions from $|N|\times c$ to $|P|\times c$ (where $c$ is the embedding size). This ensures that no matter how many voters there are, the input size to the encoder-decoder always stays the same. Since the number of projects $P$ is typically much smaller than the number of voters $N$, it allows the use of much larger PB instances (handing instances with tens of thousands of voters rather than thousands before running out of memory). In addition, we start with the projects' space from the encoder to find the connections between the projects directly instead of passing through voters' latent space.


\subsection{Training the NN}
We propose two modes of training the NN to learn an aggregation rule $R$: \emph{implicit} and \emph{explicit}. In the implicit mode, pairs $(X,B_{opt})$ are fed to the NN ($X$ is a PB instance, and $B_{opt}$ is an optimal bundle according to the rule $R$). The network computes a bundle $B$ given $X$ at the input layer (in the manner described at the end of Section~\ref{sec:arch}), and the cross entropy loss function $H(B,B_{opt})$ is used to adjust the weights. 

We have noticed empirically that the implicit approach may struggle when the PB instance has many tied optimal solutions (we discuss this in more detail in Section \ref{sec:ties}). Hence, we offer an \emph{explicit} approach for learning the aggregation rule by directly encoding the rule's score in the loss function. We considered two scores:
\begin{itemize}
    \item Social welfare: Let $s$ be the optimal social welfare achievable in the given PB; the loss at a given output $B$ is  $\ell_{SW}(B) = |1 - \frac{\sum_{i\in N} A_i \cdot B}{s}|.$
    If the budget is not respected, the welfare at $B$ can exceed $s$. To handle this, we took the absolute value of the difference. 
    \item Representation: Note that the representation score is not differentiable; therefore, we need to approximate it. Let $r$ be the optimal representation ratio achievable for a given PB and let $\sigma(x) = \frac{1}{1 + \exp^{a-bx}}$ be a shifted sigmoid; the loss at a given output $B$ is $\ell_{REP}(B)=|1 - \frac{\sum \sigma(A_i \cdot B)}{r}|$. We chose $a=3; b=6$ after testing a few candidates.

\end{itemize}

Each of the encoder/decoder parts is built on transformers with 20 heads of size 28, and to train it, Adam optimizer with a learning rate of $0.0002$ was used (similar to \citet{anil2021learning}).
The code is based on PyTorch and ran on Gtx 1090 GPU for seven days in the training phase.


\section{Datasets}\label{sec:data}

In this section, we describe the datasets that were used to train and test the networks ST and ST+PMA. 
The training datasets consist of synthetically generated PB instances, similar to datasets used in the literature as benchmarks for PB and multi-winner settings. The test datasets includes in addition to synthetically generated datasets, real-world PB instances. Table \ref{tab:amount} summarizes the statistics of each dataset.

\paragraph{Training Datasets}
In what follows we use the notation $X\sim[a,b]$ to refer to a random variable $X$ that is uniformly distributed over the interval $[a,b]$.

{\bf Euclidean (EUC)} ~\citep{enelow1984spatial,skowron2020participatory,elkind2017multiwinner,talmon2019framework}. This dataset consists of PB instances generated according to the 2-dimensional Euclidean model. Each instance is generated by sampling locations for each of the voters and projects in the $[0,1] \times [0,1]$ square; then, each voter $v$ approves $k_v$ projects that are closest to his location. There are various ways to randomize the number of voters, projects, and the $k_v$'s. For the sake of fluent reading, we put all the details in Appendix~\ref{app:euc}.

{\bf One vs. Many (OVM)}~\citep{sarkar1992simple} The instances in this set test the capacity of the network to decide between two cases, a single project vs multiple projects. The optimal choice depends on the aggregation rules and the parameters that we'll describe now. Each PB instance consists of $p \in [2,5]$ projects, all with the same cost and together consume the entire budget; each project is approved by 100 voters sampled randomly from a pool of $100p$ voters. In addition, there is another project that, if chosen, no other project can be funded; this project has $v$ fresh voters, where $v$ is a parameter of the specific PB instance. 

The next two datasets, taken from~\citet{fairstein2022welfare}, 
pose three challenges: (a)  good ratios of welfare and representation cannot be obtained simultaneously (unlike the EUC dataset for example; (b) they consist of instances for which there are many possible optimal outcomes with respect to representation (ties). This may be challenging to the implicit learning approach as only a single outcome is selected as the "ground truth"; (c) a random solution already produces a relatively high representation ratio (see Table \ref{tab:rep}, top part). 

{\bf Trade-Off AV (TOAV).} In this dataset we have $mx + x' $ voters and $m^2$ projects such that exactly $m$ projects can be funded. The voters are split into $m$ groups (each of size $x$, except the first of size $x + x'$), each approving $m$ projects without intersection. we randomize $m\in [3,7]$ (total of 9-49 projects), $x\in[5,20]$ and $x'\in [1,100]$. The optimal SW, $m(x+x')$, is obtained by funding $m$ projects from the largest group. However, this solution has a relatively low representation ($x+x'$). On the other hand, a solution that funds one project from each group (including the largest) has the best representation of $mx+x'$, but that's also the SW, which is much smaller than the optimum. Note that many solutions obtain this best representation, i.e., many ties.

{\bf Trade-Off CC (TOCC)} In this dataset we have $n_1 + n_2$ voters and $2n_1$ projects with a budget of $L$. The first $n_2$ voters approve the first $n_1$ projects, each costs $\frac{L}{n_1}$, and the rest of the voters approve a single project that costs $\frac{L}{n_1}$. We randomize $n_1\in [3,24]$ and $n_2\in [n_1, 2n_1]$. The best representation is obtained by choosing the single project and $n_1-1$ projects from the first batch, giving a representation of $n_1+n_2$ and SW of $n_2\cdot(n_1-1) + n_1 = n_2n_1 - n_2+n_1$. Note that there are $n_1$ such optimal solutions.
The maximal SW, $n_2n_1$, is obtained by choosing all first $n_1$ projects, but the representation is only $n_2$ in this case. As the gap between $n_2$ and $n_1$ increases, the trade-off becomes less favorable.

For the training stage, 30,000 instances were taken from the EUC dataset and 10,000 instances from each other. More instances were taken from EUC because it includes many sub-distributions (different choices for randomization over the choice of users, projects, and locations).

\paragraph{Testing Datasets}
We used two types of datasets to test the networks. First,  variations of  PB instances from the training set but with a larger number of voters (see the bottom line of Table~\ref{tab:amount}).
Second,  real-world participatory budgeting instances from Poland. The Poland dataset is taken from \url{Pabulib.org}~\citep{stolicki2020pabulib}, a library of PB instances available to the research community.
We tested all 130 PB instances that took place in different districts of  Warsaw, Poland, in the years 2017--2021. Each instance included between 50 to 10,000 voters (2,982 on average) and between 20 to 50 projects (36 on average).\footnote{We selected a maximum of 50 projects, as we believe that exceeding this number could impose excessive cognitive strain on voters, potentially leading to less accurate voting outcomes. However, if desired, we can increase the number of projects beyond 50 by adjusting a network parameter, without compromising the quality of the network.}

We will use the notation Dataset-Name\_size to mention to one of the sizes described in Table~\ref{tab:amount} where needed, for example EUC\_large to indicate the EUC dataset in the sampled according to the large settings.

\begin{table*}[]
\begin{center}
\begin{tabular}{c|c|c|c|c|c|c|c|c}
Dataset &\multicolumn{4}{c|}{EUC}                                          &   OVM             &     TOAV          &     TOCC    & Poland    \\
\hline
& small        & medium        & large          & xlarge           &             &            &     &     \\
\hline
Train & {[}20,100{]} & - & - & - & {[}203,909{]} & {[}16,240{]} & {[}9,120{]} & -\\ 
Test&
{[}20,100{]} & {[}100,500{]} & {[}500,1k{]} & {[}1k,10k{]} & {[}61,50k{]} & {[}66,1k{]} & {[}6,72{]} & {[}50,10000{]} 
\end{tabular}\caption{The range of number of voters in PB instances of each dataset.}\label{tab:amount}
\end{center}
\end{table*}

\subsection{Dealing with Ties}\label{sec:ties}
One aspect widely recognized in the field of PB and voting, in general, pertains to the occurrence of ties (the existence of multiple optimal solutions)~\cite{janeczko2023ties, xia2021likely}. Breaking ties can cause fairness issues and result in sub-optimal choices regarding a secondary objective. Training an NN on tied instances may also pose a challenge since only one solution has been chosen as the right ground truth.

We review the aforementioned datasets in light of the ties discussion and report, per aggregation rule, the average amount of ties. We used either an exact solver or a rigorous analysis to compute the number of ties. 

\begin{itemize}
    \item AV: For most instances (across all datasets), there is one optimal solution and, in rare cases, up to a few ties. The average of EUC\_small is 1.28 ties per instance of, and this average grows smaller as the number of voters increases.

    \item PAV: For the EUC, OVM and Poland datasets PAV rule has a single outcome except for a few instances with two outcomes. In contrast, for both TOAV and TOCC datasets, instances have up to a few dozen optimal outcomes.

    \item CC: This is the most relevant rule to consider with respect to multiple ties, as representation is much less sensitive to changes in the outcome. 
    \begin{itemize}
        \item For the EUC datasets, as the number of voters increases, the average number of ties decreases: EUC\_small has on average ~33.7 possible outcomes, ~18.2 for EUC\_medium, 14.7 for EUC\_large and ~2 for EUC\_very\_large.
        \item In OVM, there is always only one solution.
        \item In Poland there are on average 1.3 optimal soultions.
        \item For both TOAV and TOCC, there is a simple optimal solution where a single project from each "group" is chosen. However, there are many such groups. While TOCC can have up to ~1000 ties, TOAV can reach up to $~10^8$ ties in the train set and $~10^{16}$ ties in the test set.
    \end{itemize} 
\end{itemize}

Subsequently, we will categorize the datasets into two groups: the UNIQUE datasets (EUC, OVM, and Poland) with a small number of ties and the TIED datasets (TOAV and TOCC) with a significant number of ties. In the next section, we evaluate the learning capacity for UNIQUE vs TIED.


\section{Evaluation - Learning Target Distributions}\label{sec:res1}
In this section, we analyze the capacity of the different models to learn a variety of aggregation rules (representing our chosen distribution). We have four models depending on the choice of (a) learning type explicit/implicit and (b) model architecture, ST with PMA or without. In addition, we have Random, a model which picks projects at random until the budget is consumed and sequential variants of the rules (iteratively taking the project with the highest marginal score). For brevity, we don't report the performance of the ST model with explicit learning since its performance was like that of Random. 

\subsection{Evaluation Metric}

As mentioned in Section~\ref{sec:method}, to evaluate the quality of the learnt rule we would like to use social welfare and representation metrics.
Since those values can be sensitive to the size and structure of instances we would like to normalize them so they will be consistent, which is done in the following manner: given some PB instance, an outcome $B\subseteq P$ and a solution with optimal welfare $B_{opt}\subseteq P$, the welfare ratio is defined by $SW(A,B)/SW(A,B_{opt})$. Similarly, given an outcome $B\subseteq P$ and an outcome with optimal representation $B_{opt}\subseteq P$, the representation ratio is defined by $REP(A,B)/REP(A,B_{opt})$.

Once we have those values, we can consider the learnt rule distribution in the welfare-representation space instead of the Euclidean space~\citep{faliszewski2022optimization}. The idea is that given instances each rule or objective will have points in the welfare-representation space and we can compare the distribution of our target and the learnt rule. In our case we have 11 datasets we evaluate the rules, for each one we calculate the average welfare and representation achieved i.e. we get for each rule 11 points that represents the rule distribution. Then we compute the distance between our optimal and learnt rules using root mean square error (RMSE); a lower RMSE indicates a better prediction.

We note that while we chose a metric that best fit to our domain, in the ML literature a common metric to use is Jaccard Similarity. For the interested reader, we refer to the Appendix where we present those results.

\subsection{Results}

\begin{table*}[h!]
\begin{center}
\begin{tabular}{c|ccc|ccc|ccc|cc}
                     & \multicolumn{3}{c}{sequential}    & \multicolumn{3}{|c}{ST (implicit)} & \multicolumn{3}{|c}{ST+PMA   (implicit)} & \multicolumn{2}{|c}{ST+PMA   (explicit)} \\
\hline
Optimization score          & AV     & CC    & PAV & AV     & CC    & PAV   & AV           & CC          & PAV         & AV                 & CC                \\
\hline
Welfare Ratio RMSE        & 0.165 & 0.135   &   0.173   & 0.540  & 0.460 & 0.528 & 0.011        & 0.053       & 0.016       & 0.018              & 0.037             \\
Representation Ratio RMSE & 0.076 & 0.032   &   0.032   & 0.481  & 0.536 & 0.514 & 0.002        & 0.024       & 0.021       & 0.013              & 0.039 \\
\hline
Total RMSE                & 0.242  &   0.168   &   0.204   & 1.02  & 0.995 & 1.04 & \textbf{0.013}        & \textbf{0.076}       & \textbf{0.038}       & 0.032              & \textbf{0.076} 
\end{tabular}

\caption{UNIQUE results: The RMSE between the ground truth and the predictions (bold values for best total RMSE).}\label{tab:distance_unique}
\end{center}
\end{table*}

\begin{table*}[h!]
\begin{center}
\begin{tabular}{c|ccc|ccc|ccc|cc}
                     & \multicolumn{3}{c}{sequential}    & \multicolumn{3}{|c}{ST (implicit)} & \multicolumn{3}{|c}{ST+PMA   (implicit)} & \multicolumn{2}{|c}{ST+PMA   (explicit)} \\
\hline
Optimization score             & AV     & CC    & PAV & AV     & CC    & PAV   & AV           & CC          & PAV         & AV                 & CC                \\
\hline
Welfare Ratio RMSE        & 0.006 & 0   &   0.010   & 0.009  & 0.028 & 0.045 & 0.006        & 0.030       & 0.034       & 0.010              & 0.284             \\
Representation Ratio RMSE & 0.002 & 0   &   0.005   & 0.033  & 0.244 & 0.153 & 0.036        & 0.281       & 0.127       & 0.094              & 0.171 \\
\hline
Total RMSE                & 0.009  &   0   &   0.014   & 0.042  & \textbf{0.270} & 0.199 & \textbf{0.042}        & 0.312       & \textbf{0.161}       & 0.105              & 0.456 
\end{tabular}

\caption{TIED results: The RMSE between the ground truth and the predictions (bold values for best total RMSE).}\label{tab:distance_tied}
\end{center}
\end{table*}

Tables~\ref{tab:distance_unique} and \ref{tab:distance_tied} summarize the RMSE results. Detailed results, on the dataset level are included in Appendix~\ref{app:results}. In those tables the results are split to the UNIQUE and TIED datasets. The reason for this is that ML methods are known to have difficult time to learn in the case when there are many possible ties. This separation let us better understand the results and the cases where each method is preferred.

\emph{UNIQUE Datasets}.
As evident from Table~\ref{tab:distance_unique}, our approach succeed to achieve the best results. It is conspicuous the extreme different between ST and ST+PMA. We see that the addition of PMA layer improves significantly the results. Furthermore, as Described in Section~\ref{sec:method}, ST would need to handle larger inputs which exhausts the memory for the larger instances i.e. ST exhaust the memory and crash on the datasets EUC\_xlarge, OVM\_large and Poland, which means it is not capable to run on real-life instances.

We note that the RMSE for our approach is low in general, pointing to good learning and generalization capacity. Moreoever, we see there is no significant difference between explicit and implicit learning, which is not the case for the TIED dataset we now explore.

\emph{TIED Datasets}
Looking at Table~\ref{tab:distance_tied}, the most noticeable thing is that the sequential variants mimic the rules almost perfectly. This result is due to how those datasets are created, where the optimal outcome is clear and easy to find both greedily and by eyeballing the PB instance. 

Next, we compare the implicit and explicit methods. While the combined RMSE of the implicit learning is lower for both AV and CC, the representation RMSE, when optimizing for CC, is considerably better for the explicit method (0.171 vs 0.281, 40\% lower). This exhibits the main difference between the two approaches. While the explicit approach optimize directly the representation, the implicit approach optimize better the distribution called CC. Even though that CC is known to optimize the representation the way it does it have an affect on the welfare which the implicit approach take into account when learning from the examples. In contrast, the explicit approach focus only optimizing the representation, ignoring the fact that the notion of welfare exists. This motivating the choice of whether to use explicit or implicit learning according to the type of data and the goals of the PB designers.


\section{Evaluation - Learning Mixture Distributions}\label{sec:AV+CC}

In this section, we aim to elucidate the framework's ability to learn aggregation rules from examples that do not conform to a single optimal distribution. For example, these examples may originate from a committee characterized by divergent perspectives, requiring the creation of an aggregation rule based on ``collective intuition" to accommodate varied viewpoints.

To demonstrate this capability, we consider a hypothetical committee comprising two factions: those who support the maximization of social welfare, denoted as AV aggregation supporters, and those who support the maximization of representation, denoted as CC aggregation supporters. This scenario is parameterized with $p \in [0,1]$, the fraction of committee members who endorse AV outcomes. We call AV-CC-p the learned "compromise" rule.

A training dataset of size $n=30k$ for learning AV-CC-p is constructed from the UNIQUE datasets outlined in Section \ref{sec:data} as follows. We select $pn$ random instances and associate them with each AV outcome and $(1-p)n$ random instances associated with CC outcomes. We sample with repetition so the same instance can appear once with an AV solution and once with CC. The intuition behind it, given each example instance we sample one person uniformly to determine the optimal distribution, thus expression all objectives in the dataset used for learning.

Alongside AC-CC-p, we solve an ILP whose objective function is the weighted sum of welfare and representation scores,  $p \cdot welfare + (1-p) \cdot representation$. Unlike the pure $p=0$ or $p=1$ cases, we do not expect AC-CC-p to converge to the ILP 
because AC-CC-p learns from examples of two "opinions" which creates a compromise, unlike learning directly from the ILP outcomes which have a single "correct" outcome to each instance.

The evaluation of AV-CC-p and the ILP is tricky since we can't compare the selected projects to the committee's selection. Instead, we run AV-CC-p and the ILP on the UNIQUE test datasets in Section \ref{sec:data}. We compute the representation and social welfare ratio between the solution of the compromise rule and the optimal values. 

\begin{figure}[ht!]
\begin{center}
\includegraphics[width=9cm]{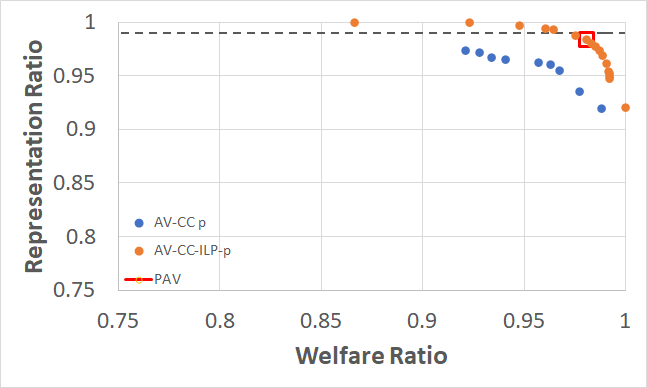}
\caption{Welfare and rep. ratios on the UNIQUE datasets for the AV-CC-p models (blue dots) and the weighted AV-CC model (orange dots). The PAV value is added as well (red square). The first four points from the left correspond to $p=0,0.01,0.04,0.8$ and the dashed line at $y=0.99$ marks a favorable AV-CC tradeoff for $p\le 0.1$. The remaining blue and orange dots are in 0.1 steps.
}\label{fig:imitation}\vspace{2mm}
\end{center}
\end{figure}

Figure~\ref{fig:imitation} plots the ratios as we change $p \in [0,1]$. Each blue dot represents the average ratio over the entire test set. Similarly, orange dots represent the ILP ratio.

As the figure shows, the blue and the orange dots lie on a curved line. The leftmost orange dot corresponds to CC ($p=0$) and the rightmost to AV ($p=1$). Traveling on the curve from CC (leftmost orange dot) to AV (rightmost), we see the welfare-representation tradeoff, but not linearly. In particular, we see for the orange dots a favorable trade-off for the first four dots ($p \le 0.1$). In that case, welfare improves by as much as 10\% while losing 1\% in representation. The blue dots follow the orange quite closely, although the zoomed-in scaling of the axes can blur this observation.

Figure~\ref{fig:imitation} also features the average optimal PAV value. It coincides with the $p \cdot welfare + (1-p)\cdot representation$ orange dot for $p=0.3$ (30\% AV, 70\% CC). It is also $p=0.3$ for the closest blue dot to PAV. This gives, for the first time, both a data-driven rationale for PAV, as well as a closed-formula one.~\footnote{When repeating the experiment for multi-winner we see similar convergence at $p=0.3$.} It is important to mention that while this AV-CC-0.3 achieves very close welfare and representation ratios on average to those of PAV, this does not mean that those two rules give the same outcome as PAV.

This section supports our hypothesis that learning new aggregation rules from examples from multiple sources is possible, where even the same instance can have different ground truths (fitting different opinions).

\section{Conclusions and Future Work}\label{sec:conc}

This work focuses on the concept of learning new aggregation rules for participatory budgeting (PB) settings from arbitrary distributions. This required the generalization of different techniques as introduced in Section~\ref{sec:method}. We start by describing arbitrary distributions through examples, defining a general metric that can be used for multi-winner and PB instances, and using set transformer architecture to better learn complex rules and create a framework for learning and evaluating new rules.


Our tests confirm that the pipeline we introduced allow learning with high accuracy interesting distributions to create new aggregation rules. Moreover, a side effect from using ML is having a very quick aggregation rule which can be critical if it being used in real-time applications such as using  multi-winner voting in real-time strategy games~\cite{jiao2016diversity} and   applications in wireless sensor networks~\cite{wu2022multiwinner}. More details about it can be found in the Appendix.


We introduced two approaches to learning aggregation methods: an explicit approach that maximizes the social criteria we wish to optimize and an implicit approach that learns directly from examples. We evaluated the approaches on varied synthetic PB datasets and real-world PB instances. We show that there is an advantage to implicit learning as it does not require formalizing the desired distribution \ objective (instead giving examples of it) and seems to get closer results to the distribution. Using explicit learning is mostly useful in scenarios where there are many possible outcomes that optimize the objective, and we want to direct the model to the distribution instead of a specific solution.

The data-driven approach to learning aggregation rules opens a window into a new way of defining aggregation rules for PB and mult-winner instances, by simply supplying examples on the objective behavior. 
Therefore, allowing PB organizers without relevant theoretical knowledge to create a rule that they desire without requiring them to understand what distribution it represents.

Two important capabilities of our approach  include  learning from small instances and generalizing well for larger instances; and the capacity to learn an aggregation rule that finds a good compromise in case of multiple opinions (the AV-CC-p rule presented in Section \ref{sec:AV+CC}). 

While our research has yielded promising results, there are a couple of limitations. First, ML approaches are considered a black box, i.e., given preferences, we get the outcome; however, the reasoning for choosing the outcome is not accessible.
Despite this limitation, much research focuses on explaining voting outcomes~\citep{hashemi2024user,boixel2022displaying, peters2020explainable}. Therefore, one can combine our approach with one of those methods to add explainability to the outcome.

Second, even if the approach works well on average, it lacks a guarantee of its worst-case performance. For those reasons, it is crucial to exercise caution when using it for classic PB, and it should be considered a supportive tool to choose the outcome rather than an absolute outcome. Nevertheless, there are many applications that the impact of having a specific instance where the approach doesn't work well is not severe. For example, in the case of groups of friends, if they are not satisfied with the outcome, they can either retry or just take another approach.


The approach that we described in the paper can be further extended. While in our paper the main focus was on the three rules AV, CC and pav, it possible possible to use it for other objectives such creating proportional rules by either implicitly learning proportional rules or describing it explicitly. In addition, while ML methods generally considered as black box, there are ways to explain the reasoning for certain output. Finally, we would like to conduct an experiment where people express their ideal outcomes, creating a rule from their intuition and showing that such rule is capable to achieve better compromise compare to any existing rule.


%
%
%
\begin{small}
\bibliographystyle{splncs04nat}

\bibliography{ref.bib}

\begin{thebibliography}{37}
\providecommand{\natexlab}[1]{#1}
\providecommand{\url}[1]{\texttt{#1}}
\providecommand{\urlprefix}{URL }
\expandafter\ifx\csname urlstyle\endcsname\relax
  \providecommand{\doi}[1]{doi:\discretionary{}{}{}#1}\else
  \providecommand{\doi}{doi:\discretionary{}{}{}\begingroup
  \urlstyle{rm}\Url}\fi

\bibitem[{Anil and Bao(2021)}]{anil2021learning}
Anil, C., Bao, X.: Learning to elect. Advances in Neural Information Processing
  Systems \textbf{34}, 8006--8017 (2021)

\bibitem[{Aziz and Shah(2021)}]{aziz2021participatory}
Aziz, H., Shah, N.: Participatory budgeting: Models and approaches. In:
  Pathways Between Social Science and Computational Social Science, pp.
  215--236, Springer (2021)

\bibitem[{Boixel et~al.(2022)Boixel, Endriss, Nardi
  et~al.}]{boixel2022displaying}
Boixel, A., Endriss, U., Nardi, O., et~al.: Displaying justifications for
  collective decisions. In: IJCAI, pp. 5892--5895 (2022)

\bibitem[{Cabannes(2004)}]{cabannes2004participatory}
Cabannes, Y.: Participatory budgeting: a significant contribution to
  participatory democracy. Environment and urbanization \textbf{16}(1), 27--46
  (2004)

\bibitem[{Caragiannis and Fehrs(2022)}]{caragiannis2022complexity}
Caragiannis, I., Fehrs, K.: The complexity of learning approval-based
  multiwinner voting rules. In: Proceedings of the AAAI Conference on
  Artificial Intelligence, vol.~36, pp. 4925--4932 (2022)

\bibitem[{Chamberlin and Courant(1983)}]{chamberlin1983representative}
Chamberlin, J.R., Courant, P.N.: Representative deliberations and
  representative decisions: Proportional representation and the borda rule.
  American Political Science Review \textbf{77}(3), 718--733 (1983)

\bibitem[{Doucette et~al.(2015)Doucette, Larson, and
  Cohen}]{doucette2015conventional}
Doucette, J.A., Larson, K., Cohen, R.: Conventional machine learning for social
  choice. In: Twenty-Ninth AAAI Conference on Artificial Intelligence (2015)

\bibitem[{Elkind et~al.(2017)Elkind, Faliszewski, Laslier, Skowron, Slinko, and
  Talmon}]{elkind2017multiwinner}
Elkind, E., Faliszewski, P., Laslier, J.F., Skowron, P., Slinko, A., Talmon,
  N.: What do multiwinner voting rules do? an experiment over the
  two-dimensional euclidean domain. In: Proceedings of the AAAI Conference on
  Artificial Intelligence, vol.~31 (2017)

\bibitem[{Enelow and Hinich(1984)}]{enelow1984spatial}
Enelow, J.M., Hinich, M.J.: The spatial theory of voting: An introduction. CUP
  Archive (1984)

\bibitem[{Fairstein et~al.(2021)Fairstein, Meir, and
  Gal}]{fairstein2021proportional}
Fairstein, R., Meir, R., Gal, K.: Proportional participatory budgeting with
  substitute projects. arXiv preprint arXiv:2106.05360  (2021)

\bibitem[{Fairstein et~al.(2022)Fairstein, Reshef, Vilenchik, and
  Gal}]{fairstein2022welfare}
Fairstein, R., Reshef, M., Vilenchik, D., Gal, K.: Welfare vs. representation
  in participatory budgeting. In: Proceedings of the 22th AAMAS, p. 409–417
  (2022)

\bibitem[{Faliszewski et~al.(2022)Faliszewski, Szufa, and
  Talmon}]{faliszewski2022optimization}
Faliszewski, P., Szufa, S., Talmon, N.: Optimization-based voting rule design:
  The closer to utopia the better. Collective Decisions: Theory, Algorithms And
  Decision Support Systems pp. 17--51 (2022)

\bibitem[{Firebanks-Quevedo(2020)}]{firebanks2020machine}
Firebanks-Quevedo, D.: Machine Learning? In MY Election? It's More Likely Than
  You Think: Voting Rules via Neural Networks. Ph.D. thesis, Oberlin College
  (2020)

\bibitem[{Hashemi et~al.(2024)Hashemi, Darejeh, and Cruz}]{hashemi2024user}
Hashemi, M., Darejeh, A., Cruz, F.: A user-centric exploration of axiomatic
  explainable ai in participatory budgeting. In: Companion of the 2024 on ACM
  International Joint Conference on Pervasive and Ubiquitous Computing, pp.
  126--130 (2024)

\bibitem[{Jain et~al.(2020)Jain, Sornat, and Talmon}]{jain2020participatory}
Jain, P., Sornat, K., Talmon, N.: Participatory budgeting with project
  interactions. In: Proceedings of the 29th International Joint Conference on
  Artificial Intelligence (IJCAI), pp. 386--392 (2020)

\bibitem[{Janeczko and Faliszewski(2023)}]{janeczko2023ties}
Janeczko, {\L}., Faliszewski, P.: Ties in multiwinner approval voting. arXiv
  preprint arXiv:2305.01769  (2023)

\bibitem[{Jiao et~al.(2016)Jiao, Xu, and Sun}]{jiao2016diversity}
Jiao, P., Xu, K., Sun, L.: Diversity voting and its application in real-time
  strategy games multi-objective optimization decision-making behavior
  modeling. In: 2016 3rd International Conference on Systems and Informatics
  (ICSAI), pp. 552--559, IEEE (2016)

\bibitem[{Kujawska et~al.(2020)Kujawska, Slavkovik, and
  R{\"u}ckmann}]{kujawska2020predicting}
Kujawska, H., Slavkovik, M., R{\"u}ckmann, J.J.: Predicting the winners of
  borda, kemeny and dodgson elections with supervised machine learning. In:
  Multi-Agent Systems and Agreement Technologies, pp. 440--458, Springer (2020)

\bibitem[{Lee et~al.(2019)Lee, Lee, Kim, Kosiorek, Choi, and Teh}]{lee2019set}
Lee, J., Lee, Y., Kim, J., Kosiorek, A., Choi, S., Teh, Y.W.: Set transformer:
  A framework for attention-based permutation-invariant neural networks. In:
  International conference on machine learning, pp. 3744--3753, PMLR (2019)

\bibitem[{Lu et~al.(2024)Lu, Peters, Aziz, Bei, and
  Suksompong}]{lu2024approval}
Lu, X., Peters, J., Aziz, H., Bei, X., Suksompong, W.: Approval-based voting
  with mixed goods. Social Choice and Welfare pp. 1--35 (2024)

\bibitem[{Orsted et~al.(1894)Orsted, Forchhammer, and
  Steenstrup}]{orsted1894oversigt}
Orsted, H.C., Forchhammer, G., Steenstrup, J.J.S.: Oversigt over det Kongelige
  Danske Videnskabernes Selskabs Forhandlinger (1894)

\bibitem[{Peters et~al.(2021)Peters, Pierczy{\'n}ski, and
  Skowron}]{peters2021proportional}
Peters, D., Pierczy{\'n}ski, G., Skowron, P.: Proportional participatory
  budgeting with additive utilities. Advances in Neural Information Processing
  Systems \textbf{34}, 12726--12737 (2021)

\bibitem[{Peters et~al.(2020)Peters, Procaccia, Psomas, and
  Zhou}]{peters2020explainable}
Peters, D., Procaccia, A.D., Psomas, A., Zhou, Z.: Explainable voting. Advances
  in Neural Information Processing Systems \textbf{33}, 1525--1534 (2020)

\bibitem[{Procaccia et~al.(2009)Procaccia, Zohar, Peleg, and
  Rosenschein}]{procaccia2009learnability}
Procaccia, A.D., Zohar, A., Peleg, Y., Rosenschein, J.S.: The learnability of
  voting rules. Artificial Intelligence \textbf{173}(12-13), 1133--1149 (2009)

\bibitem[{Rey and Maly(2023)}]{rey2023computational}
Rey, S., Maly, J.: The (computational) social choice take on indivisible
  participatory budgeting. arXiv preprint arXiv:2303.00621  (2023)

\bibitem[{Sarkar et~al.(1992)Sarkar, Chakrabarti, Ghose, and
  De~Sarkar}]{sarkar1992simple}
Sarkar, U., Chakrabarti, P., Ghose, S., De~Sarkar, S.: A simple 0.5-bounded
  greedy algorithm for the 0/1 knapsack problem. Information Processing Letters
  \textbf{42}(3), 173--177 (1992)

\bibitem[{Scarselli et~al.(2008)Scarselli, Gori, Tsoi, Hagenbuchner, and
  Monfardini}]{scarselli2008graph}
Scarselli, F., Gori, M., Tsoi, A.C., Hagenbuchner, M., Monfardini, G.: The
  graph neural network model. IEEE transactions on neural networks
  \textbf{20}(1), 61--80 (2008)

\bibitem[{Skowron et~al.(2020)Skowron, Slinko, Szufa, and
  Talmon}]{skowron2020participatory}
Skowron, P., Slinko, A., Szufa, S., Talmon, N.: Participatory budgeting with
  cumulative votes. arXiv preprint arXiv:2009.02690  (2020)

\bibitem[{Stewart et~al.(2014)Stewart, Miller, Hildreth, and
  Wright-Phillips}]{stewart2014participatory}
Stewart, L.M., Miller, S.A., Hildreth, R., Wright-Phillips, M.V.: Participatory
  budgeting in the united states: a preliminary analysis of chicago's 49th ward
  experiment. New Political Science \textbf{36}(2), 193--218 (2014)

\bibitem[{Stolicki et~al.(2020)Stolicki, Szufa, and
  Talmon}]{stolicki2020pabulib}
Stolicki, D., Szufa, S., Talmon, N.: Pabulib: A participatory budgeting
  library. arXiv preprint arXiv:2012.06539  (2020)

\bibitem[{Su(2017)}]{su2017porto}
Su, C.: From porto alegre to new york city: Participatory budgeting and
  democracy. New Political Science \textbf{39}(1), 67--75 (2017)

\bibitem[{Talmon and Faliszewski(2019)}]{talmon2019framework}
Talmon, N., Faliszewski, P.: A framework for approval-based budgeting methods.
  In: Proceedings of the AAAI Conference on Artificial Intelligence, vol.~33,
  pp. 2181--2188 (2019)

\bibitem[{Wu et~al.(2022)Wu, Chen, Zhong, Zhu, Chen, Zhang
  et~al.}]{wu2022multiwinner}
Wu, X., Chen, Z., Zhong, Y., Zhu, H., Chen, X., Zhang, P., et~al.: Multiwinner
  voting for energy-efficient mobile sink rendezvous selection in wireless
  sensor network. Wireless Communications and Mobile Computing \textbf{2022}
  (2022)

\bibitem[{Xia(2021)}]{xia2021likely}
Xia, L.: How likely are large elections tied? In: Proceedings of the 22nd ACM
  Conference on Economics and Computation, pp. 884--885 (2021)

\bibitem[{Xu et~al.(2018)Xu, Hu, Leskovec, and Jegelka}]{xu2018powerful}
Xu, K., Hu, W., Leskovec, J., Jegelka, S.: How powerful are graph neural
  networks? arXiv preprint arXiv:1810.00826  (2018)

\bibitem[{Zaheer et~al.(2017)Zaheer, Kottur, Ravanbakhsh, Poczos,
  Salakhutdinov, and Smola}]{zaheer2017deep}
Zaheer, M., Kottur, S., Ravanbakhsh, S., Poczos, B., Salakhutdinov, R.R.,
  Smola, A.J.: Deep sets. Advances in neural information processing systems
  \textbf{30} (2017)

\bibitem[{Zepic et~al.(2017)Zepic, Dapp, and Krcmar}]{zepic2017participatory}
Zepic, R., Dapp, M., Krcmar, H.: Participatory budgeting without participants:
  Identifying barriers on accessibility and usage of german participatory
  budgeting. In: 2017 Conference for E-Democracy and Open Government (CeDEM),
  pp. 26--35, IEEE (2017)

\end{thebibliography}
\end{small}

\clearpage
\begin{subappendices}


\section{Euclidean Dataset}\label{app:euc}
This section focus on the Euclidean dataset which consists PB instances generated according to the 2-dimensional Euclidean model~\citep{enelow1984spatial,skowron2020participatory,elkind2017multiwinner,talmon2019framework}, giving a detailed explanation on how the distribution for the Euclidean dataset is created.

The following properties were sampled to create the instances:
\begin{enumerate}
    \item Number of voters.
    \item Number of projects.
    \item Budget.
    \item Projects cost
    \item Voters and projects coordinates (The location on a $[0,1]X[0,1]$ map of each voter and project).
    \item Number of projects each voter approves ($k$).
\end{enumerate}

The first three items are sampled uniformly, such that the number of voters is in range according to the dataset size as described in Table~\ref{tab:amount}, the number of projects in range $[20,50]$ and the budget in range $[10000,250000]$.

{\bf Projects cost} Three types of distributions were chosen to simulate different situations (each instance have equal probability to be from each distribution):
\begin{enumerate}
    \item All costs are possible equally - the costs are sampled for uniform distribution in range $[100,100000]$.
    
    \item There are many cheap projects and a few expensive - for this purpose an exponential distribution is used, where the minimal project cost is taken at random from $(10, 20, 500, 1000)$ and the projects mean cost is taken at random from $(10000, 15000, 30000)$.
    
    \item There are few projects that are cheap or expensive - a normal distribution is used $N(\frac{budget}{2},\frac{budget}{5})$.
\end{enumerate}

{\bf Voters and projects coordinates} Two types of distributions were chosen to simulate different situations (each instance have equal probability to be from each distribution):
\begin{enumerate}
    \item Spreading equally in the entire map - the location is sampled uniformly in the entire map.
    \item More citizen or projects closer to the city center - the location is sampled using normal distribution$\sim N(0.5,sigma)$, where $sigma\in[0.1,0.2,0.3]$ taken in random (creating different city densities).
\end{enumerate}
Notice that that any combinations for the voters and projects is valid,e.g. the voters can be sampled uniformly and projects normally, or any other combination.

{\bf Number of projects each voter approves ($k$)} Two types of distributions were chosen to simulate different situations (each instance have equal probability to be from each distribution):
\begin{enumerate}
    \item It is equally likely for a voter to approve any number of projects - the number of approved projects sampled uniformly from $k_v\sim U(1,0.75|P|)$.
    
    \item It is not likely for voters approve either a few projects or most of the projects - sampled from normal distribution $k_v\sim N(\frac{|P|}{scale},3)$,  where $scale\in[3,2.5,2]$.
\end{enumerate}
In the instance creating, each voter approve the $k_v$ projects that are the closest to his location.


\section{Detailed Results}\label{app:results}

\subsection{Jaccard Similarity Results}

\begin{figure}[ht!]
\begin{center}
\includegraphics[width=9cm]{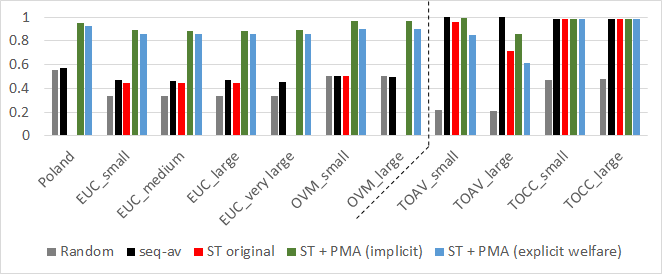}
\includegraphics[width=9cm]{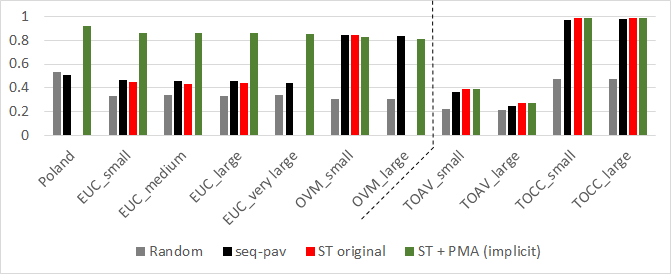}
\includegraphics[width=9cm]{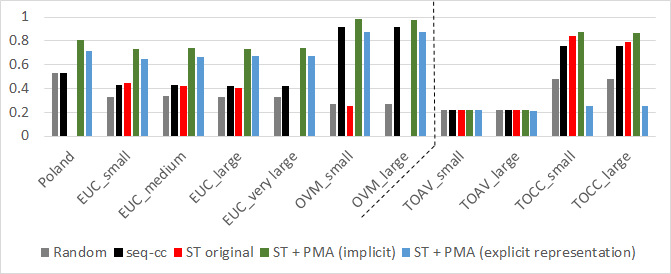}
\caption{The Jaccard similarity obtained for each of the aggregation methods: top - AV, middle - PAV , bottom - CC.  Missing values for ST is due to
GPU out-of-memory exception. UNIQUE-datasets on the left of the
dashed line and TIED-datasets on the right.
}\label{fig:full_sims}
\end{center}
\end{figure}

Figure~\ref{fig:full_sims}  show the Jaccard similarity for all sizes of the datasets. As can be seen while the NN was trained only on the small datasets, the similarity is unaffected by the instances size.

\subsection{Welfare and Representation Ratios}

Tables~\ref{tab:welfare}-\ref{tab:rep} show the welfare and representation ratios each of the methods achieved for all of the datasets.

\begin{table*}[h]
\begin{center}
\begin{tabular}{c|ccc|ccc|c|ccc}
                & \multicolumn{3}{c|}{Ground truth} & \multicolumn{3}{c|}{sequintial} & Random & \multicolumn{3}{|c}{ST} \\
\hline
dataset         & $AV$      & $CC$         & $PAV$ &    $AV$      & $CC$         & $PAV$ &         & $AV_{implicit}$     & $CC_{implicit}$    & $PAV_{implicit}$   \\
\hline
Poland          & 1       & 0.802      & 0.979 &    0.783   &   0.558   &  0.692  & 0.624     & 0      & 0     & 0     \\
EUC\_small      & 1       & 0.817      & 0.988 &    0.763   &   0.648   &   0.751 & 0.536     & 0.740  & 0.737 & 0.743 \\
EUC\_medium     & 1       & 0.840      & 0.990 &    0.755   &   0.658   &   0.744 & 0.538     & 0.735  & 0.729 & 0.729 \\
EUC\_large      & 1       & 0.856      & 0.991 &    0.755   &   0.663   &   0.744 & 0.539     & 0.735  & 0.726 & 0.734 \\
EUC\_xlarge     & 1       & 0.866      & 0.990 &    0.737   &   0.657   &   0.731 & 0.546     & 0      & 0     & 0     \\
OVM\_small      & 1       & 0.944      & 0.965 &    0.925   &   0.925   &   0.925 & 0.859     & 0.927  & 0.872 & 0.928 \\
OVM\_large      & 1       & 0.942      & 0.965 &    0.924   &   0.924   &   0.924 & 0.864     & 0      & 0     & 0     \\
TOAV\_small      & 1       & 0.703      & 0.843 &   \bf{1}       &   0.702   &   0.842 & 0.702     & 0.997  & 0.770 & 0.960 \\
TOAV\_large      & 1       & 0.835      & 0.874 &   \bf{1}       &   0.834   &   0.866 & 0.835     & 0.989  & 0.880 & 0.969 \\
TOCC\_small      & 1       & 0.165      & 1.000 &   0.984   &   0.165   &   0.974 & 0.512     & 0.985  & 0.120 & \bf{0.985} \\
TOCC\_large      & 1       & 0.145      & 1.000 &   0.984   &   0.145   &   0.980 & 0.503     & 0.985  & 0.151 & \bf{0.985}
\end{tabular}

\vspace{2em}

\begin{tabular}{c|cccccc}
                & \multicolumn{6}{c}{ST + PMA}                     \\
\hline
dataset         & $AV_{implicit}$   & $0.5AV+0.5CC_{implicit}$     & $CC_{implicit}$   & $PAV_{implicit}$   & $AV_{explicit}$ & $CC_{explicit}$ \\
\hline
Poland          & \bf{0.993} $\pm$ 0.011 & 0.984 $\pm$ 0.016 & 0.883 $\pm$ 0.159 & 0.980 $\pm$ 0.024 & 0.982 $\pm$  0.020  & 0.852 $\pm$ 0.120          \\
EUC\_small      & \bf{0.982} $\pm$ 0.074 & 0.962 $\pm$ 0.083 & 0.911 $\pm$ 0.110 & 0.974 $\pm$ 0.080 & 0.970 $\pm$  0.080  & 0.892 $\pm$    0.116       \\
EUC\_medium     & \bf{0.981} $\pm$ 0.053 & 0.964 $\pm$ 0.062 & 0.923 $\pm$ 0.086 & 0.975 $\pm$ 0.057 & 0.970 $\pm$  0.058  & 0.895 $\pm$        0.098   \\
EUC\_large      & \bf{0.983} $\pm$ 0.077 & 0.967 $\pm$ 0.083 & 0.926 $\pm$ 0.099 & 0.975 $\pm$ 0.079 & 0.972 $\pm$  0.081  & 0.897 $\pm$         0.106  \\
EUC\_xlarge     & \bf{0.982} $\pm$ 0.059 & 0.968 $\pm$ 0.068 & 0.926 $\pm$ 0.087 & 0.974 $\pm$ 0.065 & 0.971 $\pm$  0.065  & 0.896 $\pm$    0.097       \\
OVM\_small      & \bf{0.999} $\pm$ 0.013 & 0.970 $\pm$ 0.062 & 0.942 $\pm$ 0.093 & 0.932 $\pm$ 0.104 & 0.994 $\pm$  0.032  & 0.943 $\pm$ 0.091       \\
OVM\_large      & \bf{0.999} $\pm$ 0.016 & 0.969 $\pm$ 0.063 & 0.939 $\pm$ 0.096 & 0.932 $\pm$ 0.106 & 0.993 $\pm$   0.037 & 0.941 $\pm$  0.092         \\
TOAV\_small      & \bf{0.999} $\pm$ 0.002 & 0.974 $\pm$ 0.034 & 0.746 $\pm$ 0.252 & 0.924 $\pm$ 0.068 & 0.981 $\pm$   0.026 & 0.863 $\pm$       0.091    \\
TOAV\_large      & \bf{0.995} $\pm$ 0.006 & 0.976 $\pm$ 0.024 & 0.851 $\pm$ 0.212 & 0.949 $\pm$ 0.053 & 0.981 $\pm$  0.020  & 0.910 $\pm$        0.089   \\
TOCC\_small      & \bf{0.985} $\pm$ 0.029 & 0.961 $\pm$ 0.043 & 0.095 $\pm$ 0.103 & 0.982 $\pm$ 0.031 & 0.984 $\pm$   0.030 & 0.805 $\pm$    0.098       \\
TOCC\_large      & \bf{0.985} $\pm$ 0.030 & 0.971 $\pm$ 0.040 & 0.081 $\pm$ 0.122 & 0.984 $\pm$ 0.031 & 0.983 $\pm$  0.031  & 0.816 $\pm$    0.095
\end{tabular}\caption{The welfare ratio between the optimal solution and the bundle predicted by the NN. The best ratio is in bold. Zeros mean GPU out-of-memory error for all instances.}\label{tab:welfare}
\end{center}
\end{table*}

\begin{table*}[h]
\begin{center}
\begin{tabular}{c|ccc|ccc|c|ccc}
                & \multicolumn{3}{c|}{Ground truth} & \multicolumn{3}{c|}{sequintial} & Random & \multicolumn{3}{|c}{ST} \\
\hline
dataset         & $AV$      & $CC$         & $PAV$    &     $AV$      & $CC$         & $PAV$    &  & $AV_{implicit}$     & $CC_{implicit}$    & $PAV_{implicit}$   \\
\hline
Poland          & 0.940 & 1 & 0.974 &   0.924   &   0.950   &   0.946 & 0.843 & 0       & 0      & 0       \\
EUC\_small      & 0.957 & 1 & 0.980 &   0.903   &   \bf{0.960}   &  0.940 & 0.828 & 0.893   & 0.908  & 0.905   \\
EUC\_medium     & 0.963 & 1 & 0.984 &   0.901   &   0.952   &   0.933 & 0.836 & 0.894   & 0.908  & 0.904   \\
EUC\_large      & 0.966 & 1 & 0.985 &   0.901   &   0.951   &   0.931 & 0.839 & 0.897   & 0.908  & 0.909   \\
EUC\_xlarge     & 0.967 & 1 & 0.985 &   0.896   &   0.947   &   0.926 & 0.849 & 0       & 0      & 0       \\
OVM\_small      & 0.823 & 1 & 0.989 &   0.981   &   0.981   &   0.981 & 0.620 & 0.983   & 0.630  & 0.984   \\
OVM\_large      & 0.829 & 1 & 0.989 &   0.982   &   0.982   &   0.982 & 0.631 & 0       & 0      & 0       \\
TOAV\_small      & 0.358 & 1 & 0.919 &  0.357   &   \bf{1}       &   0.918 & 0.715 & 0.377   & 0.650  & 0.559   \\
TOAV\_large      & 0.277 & 1 & 0.964 &  0.276   &   \bf{1}       &   0.969 & 0.715 & 0.381   & 0.674  & 0.602   \\
TOCC\_small      & 0.619 & 1 & 0.619 &  0.625   &   \bf{1}       &   0.632 & 0.839 & 0.626   & 0.550  & 0.626   \\
TOCC\_large      & 0.715 & 1 & 0.715 &  0.720   &   \bf{1}       &   0.723 & 0.877 & 0.720   & 0.525  & 0.720 
\end{tabular}

\vspace{2em}

\begin{tabular}{c|cccccc}
                & \multicolumn{6}{c}{ST + PMA}                     \\
\hline
dataset         & $AV_{implicit}$   & $0.5AV+0.5CC_{implicit}$   & $CC_{implicit}$    & $PAV_{implicit}$   & $AV_{explicit}$ & $CC_{explicit}$ \\
\hline
Poland          & 0.935 $\pm$  0.059 & 0.933 $\pm$ 0.059 & \bf{0.964} $\pm$ 0.059 & 0.937 $\pm$ 0.054 & 0.927 $\pm$ 0.055   & 0.932 $\pm$   0.055        \\
EUC\_small      & 0.954 $\pm$ 0.072 & 0.958 $\pm$ 0.069  & \bf{0.960} $\pm$ 0.076 & 0.958 $\pm$ 0.070 & 0.953  $\pm$ 0.073  & 0.946 $\pm$    0.083       \\
EUC\_medium     & 0.962 $\pm$ 0.058 & \bf{0.963} $\pm$ 0.059  & 0.962 $\pm$ 0.066 & \bf{0.963} $\pm$  0.058 & 0.961 $\pm$ 0.059  & 0.953 $\pm$      0.071     \\
EUC\_large      & 0.965 $\pm$ 0.061 & 0.966 $\pm$ 0.060  & 0.966 $\pm$ 0.066 & \bf{0.968} $\pm$ 0.060 & 0.965 $\pm$ 0.061   & 0.956  $\pm$   0.073       \\
EUC\_xlarge     & 0.967 $\pm$ 0.054 & \bf{0.968} $\pm$ 0.056  & 0.967 $\pm$ 0.064 & \bf{0.968} $\pm$ 0.056 & 0.965 $\pm$  0.058  & 0.958 $\pm$    0.068       \\
OVM\_small      & 0.824 $\pm$ 0.249 & 0.962 $\pm$ 0.105  & \bf{0.997} $\pm$ 0.025 & 0.957 $\pm$ 0.137 & 0.850 $\pm$  0.231  & 0.961  $\pm$   0.126       \\
OVM\_large      & 0.830 $\pm$ 0.243 & 0.963 $\pm$  0.100 & \bf{0.997} $\pm$ 0.031 & 0.954 $\pm$ 0.137 & 0.855 $\pm$  0.227  & 0.963  $\pm$   0.121      \\
TOAV\_small      & 0.370 $\pm$ 0.177 & 0.502 $\pm$ 0.193  & 0.649 $\pm$ 0.197 & 0.637 $\pm$ 0.159 & 0.494 $\pm$   0.174 & 0.694 $\pm$    0.141       \\
TOAV\_large      & 0.414 $\pm$ 0.174 & 0.571 $\pm$ 0.203  & 0.681 $\pm$ 0.163 & 0.647 $\pm$ 0.170 & 0.570 $\pm$  0.198  & 0.691  $\pm$   0.143       \\
TOCC\_small      & 0.626 $\pm$ 0.055 & 0.636 $\pm$ 0.051  & 0.476 $\pm$ 0.121 & 0.626 $\pm$ 0.055 & 0.626 $\pm$  0.055  & 0.706  $\pm$   0.054       \\
TOCC\_large      & 0.720 $\pm$ 0.084 & 0.725 $\pm$ 0.079  & 0.388 $\pm$ 0.191 & 0.720 $\pm$ 0.084 & 0.721 $\pm$  0.084  & 0.779  $\pm$   0.070
\end{tabular}\caption{The ratio between the optimal solution and the bundle predicted by the NN. The best ratio is in bold. Zeros mean GPU out-of-memory error for all instances. }\label{tab:rep}
\end{center}
\end{table*}

\end{subappendices}

\end{document}